\documentclass[letterpaper]{article}
\usepackage{paper}
\usepackage[hyphens]{url}  % DO NOT CHANGE THIS
\usepackage{graphicx} % DO NOT CHANGE THIS
\usepackage{natbib}  % DO NOT CHANGE THIS AND DO NOT ADD ANY OPTIONS TO IT
\usepackage{caption} % DO NOT CHANGE THIS AND DO NOT ADD ANY OPTIONS TO IT
\usepackage{algorithm}
\usepackage{algorithmic}
\usepackage{amsmath}
\usepackage{amssymb}
\usepackage{multirow}
\usepackage{enumitem}
\usepackage[most]{tcolorbox}

\newtcolorbox{sideprompt}[1]{
  enhanced,
  colback=white,
  colframe=gray!70,
  leftrule=2pt,
  rightrule=0pt,
  toprule=0pt,
  bottomrule=0pt,
  sharp corners,
  left=10pt,
  top=12pt,
  bottom=5pt,
  fontupper=\small\itshape,
  breakable,
  overlay={
    \node[anchor=west, font=\bfseries\sffamily\tiny, color=gray!80]
    at ([xshift=10pt, yshift=-6pt]frame.north west) {#1};
  }
}

\usepackage{newfloat}
\usepackage{listings}
\DeclareCaptionStyle{ruled}{labelfont=normalfont,labelsep=colon,strut=off} % DO NOT CHANGE THIS
\floatstyle{ruled}
\newfloat{listing}{tb}{lst}{}
\floatname{listing}{Listing}

\usepackage{booktabs}

\title{EvoState: Closed-Loop Visual State Management for Long-Form Video Generation}
\author{
    Xinlei Yin\textsuperscript{\rm 1}\thanks{This work was done at Microsoft Research Asia.},
    Xiulian Peng\textsuperscript{\rm 2},
    Xiao Li\textsuperscript{\rm 2},
    Zhiwei Xiong\textsuperscript{\rm 1},
    Yan Lu\textsuperscript{\rm 2}
}
\affiliations{
    \textsuperscript{\rm 1}University of Science and Technology of China\\
    \textsuperscript{\rm 2}Microsoft Research Asia
}

\begin{document}

\maketitle

\begin{abstract}
Multi-shot long-form video generation remains challenging due to identity drift and compounding inconsistencies across shots. Existing methods encode past shots as compressed context, implicit features, or retrieved frames, making it hard to tell whether new visual evidence is another view of an existing appearance, an evolved state of the same identity, or a newly introduced entity. We propose \emph{EvoState}, an agentic framework that formulates multi-shot generation as entity-centric evolving visual state management, following a \emph{state--observation--update--propagation} cycle. An entity-centric Visual State Memory separates persistent identity anchors from their evolved states and observations, so each shot retrieves the required state, accumulates complementary views, and reactivates earlier appearances without conflating them. 
A Visual State Analyzer reconciles the text--visual--memory triplet by observing generated visual evidence, updating generation conditions according to realized outcomes and memory states, and propagating inferred state transitions into memory and future prompts. As a result, generation is conditioned on an evolving state history grounded in what has been realized, rather than solely on planned textual descriptions.
Experiments on our curated \emph{StoryBench} benchmark demonstrate substantial improvements in identity preservation, state-transition correctness, and narrative coherence over state-of-the-art methods.
\end{abstract}

\maketitle

\section{Introduction}
Recent advances have improved the visual quality, motion, and controllability of video generation~\cite{deepmind2025veo,openai2025sora2,kong2024hunyuanvideo,wan2025wan,yang2024cogvideox,kuaishou2025kling}. However, generating long-form videos with connected shots remains challenging. The model must keep characters, objects, scenes, and events consistent across changes in viewpoint and action. Small early discrepancies can compound into identity drift, wrong states, and discontinuities in later shots.

Existing methods promote multi-shot consistency in several ways. Explicit-guidance pipelines provide storyboards, layouts, or keyframes~\cite{lin2024videodirectorgpt,zhou2024storydiffusion,wang2025storyanchors,long2024videostudio,zhao2025moviedreamer,zhang2025stage}, while model-centric methods encode cross-shot dependencies through long-context attention, shot-aware features, or reference conditioning~\cite{guo2025long,meng2025holocine,wang2026multishotmaster}, leaving cross-shot dependencies implicit. Memory-augmented methods retain representative generated frames and retrieve or compress them as conditioning for later shots~\cite{zhang2025storymem,an2025onestory}. Although this strategy gives later shots direct visual evidence of prior appearances, memories organized or retrieved at the frame level do not explicitly bind evidence to entity states and may return a visually similar but temporally stale appearance. Agentic systems instead expose production as explicit stages such as script director, storyboard planner, and multimodal generator~\cite{zhuang2024vlogger,hu2024storyagent,xie2024dreamfactory,huang2026filmaster,wu2025movieagent}, and recent works introduce multimodal feedback that verifies and revises the current result~\cite{sun2026muse}. Such feedback improves alignment with current prompts through regeneration, yet it is consumed as a one-off critique instead of being accumulated into persistent states that guide later shots. 

These limitations point to the need for a generation process that treats visual history not as a pool of past frames, but as an evolving set of entity states. In a story, an entity is first established, observed from new viewpoints, may undergo appearance or state changes, and may later need to reactivate an earlier state. Frame-level memories or implicit context are often insufficient to distinguish these cases: a prompt such as ``she wears the coat again'' may retrieve the most recent or most visually similar frame rather than the specific previously established state. Separating persistent identity from evolving appearance states makes these conditions resolvable, but only if the states are continuously observed, updated, and propagated as generation proceeds.

We address this gap by formulating multi-shot video generation as entity-centric evolving visual state management organized around a \emph{state--observation--update--propagation} cycle. Our \textbf{EvoState} maintains an entity-centric \textbf{Visual State Memory} that separates identity anchors and their evolved \emph{states}, so later shots can retrieve the required state, accumulate complementary views, and reactivate prior appearances without conflating them. 
To continuously align narrative intent with realized visual outcomes, we introduce a \textbf{Visual State Analyzer} that reconciles the text--visual--memory triplet. It \emph{observes} generated keyframes and videos as visual evidence, \emph{updates} generation conditions by refining underspecified prompts and grounding planned actions in the realized scene, and \emph{propagates} inferred state transitions into memory and subsequent prompts. Through this closed-loop synchronization process, prompts, generated visuals, and memory states evolve together, improving state consistency, narrative coherence, and long-range continuity.
To precisely evaluate these capabilities, we curate \textbf{StoryBench}, a multi-shot video benchmark with controlled appearance transitions, state reactivation, delayed entity introduction, and changing viewpoints. The ablations validate the contributions of explicit entity-centric memory and the proposed text-visual-memory synergistic mechanism.

Our contributions are threefold: (1) We introduce an entity-centric Visual State Memory that separates persistent identities from evolving appearance states for explicit state management across shots. (2) We propose a Visual State Analyzer that synchronizes the text-visual-memory triplet through observation-driven state refinement and propagation, improving long-range consistency and narrative coherence. (3) We curate StoryBench, a benchmark for evaluating multi-shot video generation under controlled entity-state evolution and continuity challenges. Extensive experiments validate the effectiveness of the proposed memory formulation and the observation-driven synergistic mechanism.

\begin{figure*}[ht]
  \centering
  \includegraphics[width=0.98\linewidth]{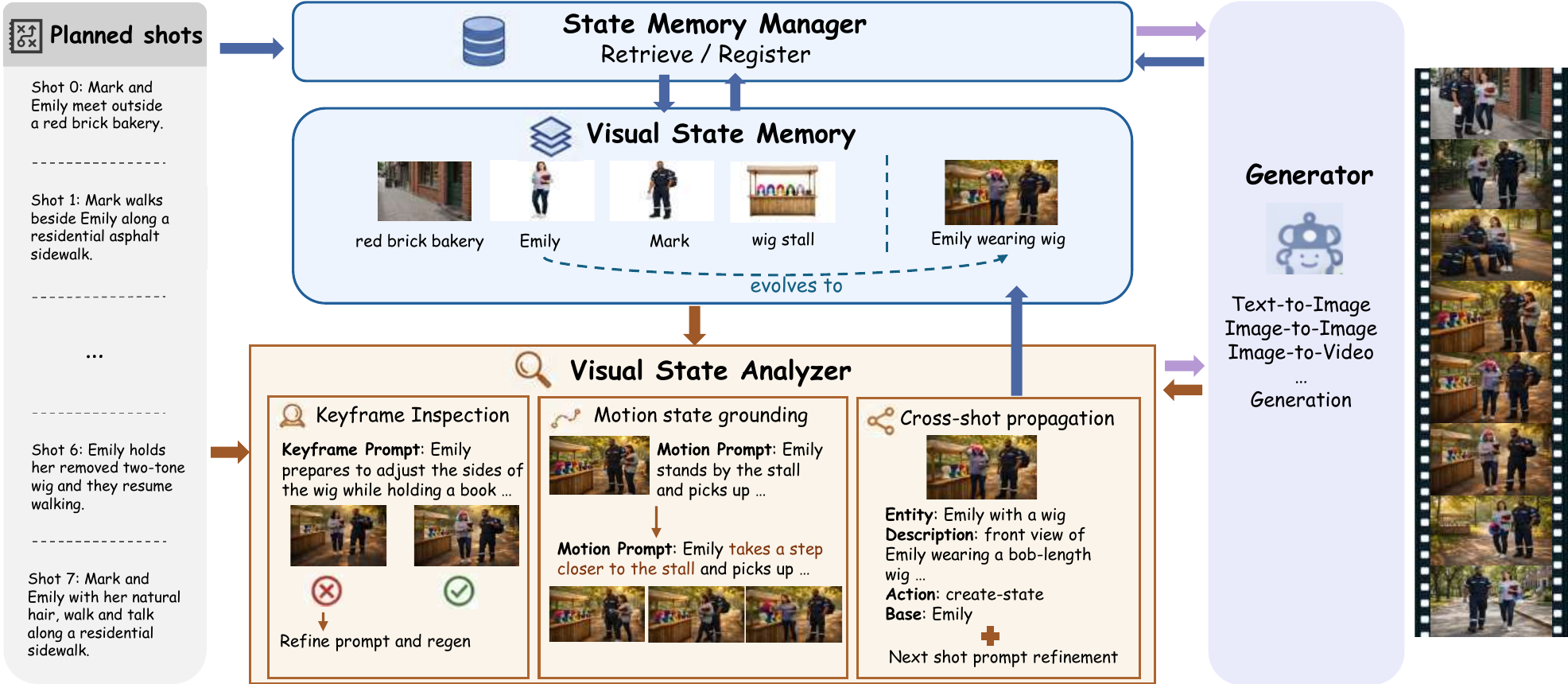}
  \caption{Overview framework. The Visual State Analyzer observes generated keyframes and video evidence to perform keyframe inspection, motion state grounding and cross-shot propagation while managing the Visual State Memory.}
  \label{fig:overview}
\end{figure*}

\section{Related works}

\subsection{Multi-Shot Video Generation}

Multi-shot generation commonly requires storyboards, keyframes, and entity references for coherent narratives~\cite{li2019storygan,rahman2023make,he2025dreamstory,wang2025characonsist,xiao2025captain,huang2024incontextlora,long2024videostudio,zhao2025moviedreamer,zheng2025videogenofthought,wang2025storyanchors}. StoryDiffusion~\cite{zhou2024storydiffusion} generates a consistent image sequence by sharing character features through consistent self-attention, then converts adjacent keyframes to video with a semantic-motion predictor. STAGE~\cite{zhang2025stage} represents a shot with a start--end storyboard pair, then retrieves a multi-shot memory pack to condition transition generation.
Other methods model inter-shot dependencies inside the generator \cite{chen2025skyreels,guo2025long,cui2026selfforcing,huang2025self,kara2025shotadapter}. HoloCine~\cite{meng2025holocine} retains only sparse inter-shot attention links while injecting localized shot prompts, balancing global narrative context against independent camera and action control; MultiShotMaster~\cite{wang2026multishotmaster} adds shot-aware positional encodings and injects grounded reference signals to control shot arrangement, subject motion, and scenes. MoC~\cite{cai2026mixture} and MoGA~\cite{jia2026moga} select and route a subset of historical tokens rather than attending over the entire history, whereas EchoShot~\cite{wang2025echoshot} embeds shot indices in a video diffusion transformer to maintain portrait identity across shots. These approaches improve context construction, but typically leave evolving entity states implicit in anchors or model features.

\subsection{Memory-Augmented Video Generation}
Recent memory-augmented multi-shot video methods condition each new shot on the visual memory constructed from previously generated shots. StoryMem~\cite{zhang2025storymem} writes semantically distinct and aesthetically reliable keyframes into a bounded memory bank, retains early anchors and recent observations through memory-sink and sliding-window policies, and injects the stored latents. OneStory~\cite{an2025onestory} encodes all prior shots as a global latent memory, queries it with the current caption to select relevant frames, and adaptively patchifies them into compact context tokens for direct conditioning. Such methods effectively retain useful history, but generally treat memory entries as frames, tokens, or compressed context rather than explicitly distinguishing identity-preserving views from temporally evolved entity states.

\subsection{Agentic Video Generation}

Agentic systems decompose production into planning and specialized roles. VideoDirectorGPT~\cite{lin2024videodirectorgpt} expands a user prompt into scenes, entities, layouts, and consistency groups for grounded generation; Vlogger~\cite{zhuang2024vlogger}, StoryAgent~\cite{hu2024storyagent}, and MovieAgent~\cite{wu2025movieagent} assign scriptwriting, directing, and storyboard-design roles to coordinate this workflow. DreamFactory~\cite{xie2024dreamfactory} and FilMaster~\cite{huang2026filmaster} similarly expose script, storyboard, and rendering stages as agent operations, while MUSE~\cite{sun2026muse} closes its multi-agent workflow with multimodal feedback that iteratively revises narrative, visual, and audio constraints.

Existing multi-shot methods improve how cross-shot context is utilized, memory-augmented approaches preserve informative history, and agentic systems expose planning, coordination, and feedback as explicit stages. We instead examine the persistent state underlying them: how evolved entity states are organized, updated from visual observations, and propagated through text and memory to guide later shots.

\section{Methodology}
\label{sec:method}

\subsection{Overview}
Given the keyframe and motion textual prompts $\{(q_t^0,m_t^0)\}_{t=1}^{T}$ for $T$ shots, where $q_t^0$ specifies the keyframe visual content of shot $t$ and $m_t^0$ defines the shot motion. These pre-specified conditions encode narrative intent but cannot fully capture realized appearance, scene structure, object state, viewpoint, or visual history. The EvoState therefore does not execute generation directly from textual prompts but maintains an entity-centric \emph{Visual State Memory} $M_t$ recording which entity states have been established up to shot $t$, resolving every planned condition $(q_t^0,m_t^0)$ against this memory into $(q_t,m_t)$ before generation, and evolving the memory from generated evidence afterwards.

Each shot thus instantiates one cycle of \emph{state--observation--update--propagation}. The State Memory Manager first resolves the \emph{states} required by $q_t$ into a reference set $R_t$. A shared vision-language-model-based (VLM) \emph{Visual State Analyzer} then realizes the observation--update--propagation in three roles: it inspects candidate keyframes against $q_t$ and $M_t$, grounds the planned motion $m_t^0$ in the accepted keyframe $I_t$ to $m_t$, and reads the resulting video $V_t$ as structured updates that are written back into $M_{t+1}$ and propagated into subsequent condition $q_{t+1}$. Implemented with three task-specific schemas (\texttt{inspect}, \texttt{ground}, and \texttt{propagate}), the Analyzer operates over a shared state representation and produces executable decisions rather than free-form critiques. Figure~\ref{fig:overview} and Algorithm~\ref{alg:vtm_loop} illustrate the process.

\begin{algorithm}[t]
\caption{Text-Visual-Memory Synergistic Generation}
\label{alg:vtm_loop}

\begin{algorithmic}[1]
\renewcommand{\algorithmicrequire}{\textbf{Input:}}
\renewcommand{\algorithmicensure}{\textbf{Output:}}

\REQUIRE Shot descriptions
$\{(q_t^0,m_t^0)\}_{t=1}^{T}$;
maximum refinement rounds $K$

\ENSURE Generated video clips $\{V_t\}_{t=1}^{T}$

\STATE $M_1 \gets \emptyset$
\COMMENT{Initialize Visual State Memory}

\STATE $q_1 \gets q_1^0$
\COMMENT{Initialize current textual prompt}

\FOR{$t=1$ \TO $T$}

    \STATE $(M_t,R_t)
    \gets
    \operatorname{ManageStateMemory}(M_t,q_t)$

    \STATE $C_t \gets \emptyset$
    \COMMENT{Candidate keyframe set}

    \STATE $I_t \gets \varnothing$
    \COMMENT{Selected keyframe}

    \FOR{$k=1$ \TO $K$}

        \STATE $I_t^{(k)}
        \gets
        \operatorname{GenerateKeyframe}(q_t,R_t)$

        \STATE $C_t
        \gets
        C_t \cup \{I_t^{(k)}\}$

        \STATE
        $(a_t^{(k)},q_t^{(k)})
        \gets
        \operatorname{Analyzer}_{\mathrm{inspect}}
        (I_t^{(k)},M_t,q_t)$

        \IF{$a_t^{(k)}=\mathrm{accept}$}
            \STATE $I_t \gets I_t^{(k)}$
            \STATE \textbf{break}
        \ENDIF

        \STATE $q_t \gets q_t^{(k)}$
        \COMMENT{Refined textual prompt}

        \STATE $(M_t,R_t)
        \gets
        \operatorname{ManageStateMemory}(M_t,q_t)$

    \ENDFOR

    \IF{$I_t=\varnothing$}
        \STATE $I_t
        \gets
        \operatorname{SelectBestCandidate}(C_t)$
    \ENDIF

    \STATE $m_t
    \gets
    \operatorname{Analyzer}_{\mathrm{ground}}
    (I_t,m_t^0)$

    \STATE $V_t
    \gets
    \operatorname{GenerateVideo}(I_t,m_t)$

    \IF{$t<T$}

        \STATE $(U_t,q_{t+1})
        \gets
        \operatorname{Analyzer}_{\mathrm{propagate}}
        (V_t,M_t,m_t,q_{t+1}^0)$

        \STATE $M_{t+1} \gets \operatorname{WriteBack}(M_t,U_t)$

    \ENDIF

\ENDFOR

\RETURN $\{V_t\}_{t=1}^{T}$

\end{algorithmic}
\end{algorithm}

\subsection{Entity-Centric Visual State Memory}
\label{sec:memory}
Past frames or compressed histories preserve rich visual evidence, but they often entangle multiple entities and their evolving appearances, making it difficult to retrieve a particular entity and associate its observations with the correct states over time. We therefore represent the persistent context as a mutable, entity-centric \emph{Visual State Memory} $M_t$ whose entries are discrete, reusable entity states, i.e.
\begin{equation}
M_t = \{e_i\}_{i=1}^{N_t}, \qquad e_i = (n_i, d_i, P_i, b_i),
\end{equation}
where $n_i$ is the entity name, $d_i$ is its textual description, $P_i$ is the set of reference images observing this state, and $b_i$ optionally links the entry to an \emph{identity anchor}. An entry with an empty $b_i$ is a base state, whereas a non-empty $b_i$ marks an evolved state or a continuity-critical view bound to the identity under which the entity first appeared. Identity therefore stays persistent while appearance evolves. For example, when Emily puts on a wig, ``Emily wearing a wig'' is registered as a derived state linked to the anchor ``Emily,'' so a later shot can request the wig state, the original state, or an additional view of either without conflating them.

\subsubsection{Pre-generation state retrieval and registration}

The memory is initialized as empty and evolves incrementally across shots as generation progresses.
Before shot $t$, the State Memory Manager retrieves from $M_t$ the states $R_t$ whose identity, appearance, and viewpoint match the current condition $q_t$ with a VLM. If a required state is missing, it invokes the Visual Generator to synthesize and register a state reference, so that generation is never conditioned on an unresolved state. The manager is re-invoked after every condition refinement, which may name a state that has not yet been established in memory.

\subsubsection{Post-generation state write-back}
After shot $t$, the Visual State Analyzer emits a set of observed state updates $U_t=\{(u_j,c_j)\}$, where each pair consists of an observed entity state $u_j$ and a structured write-back action $c_j$. The Analyzer selects one of four actions: \emph{append-view}, which adds a complementary reference to an existing state; \emph{create-state}, which registers a continuity-critical evolved state under an existing identity anchor; \emph{create-entity}, which registers a newly introduced entity; and \emph{ignore}, which discards transient, ambiguous, or continuity-irrelevant observations. The updates are then applied as
\begin{equation}
M_{t+1}=\operatorname{WriteBack}(M_t,U_t),
\end{equation}
allowing the memory to reflect realized visual state changes while avoiding fragmentation from every observed visual variation.

\subsection{Visual State Analyzer with Triplet Synergy}
\label{sec:vtm_main}
Textual prompts, stored states, and generated visual evidence are individually insufficient for maintaining long-range coherence because prompts under-specify visual details, while stored states and generated visuals may diverge from the intended narrative. The Visual State Analyzer reconciles the \textit{text-visual-memory triplet} through three operations. It performs \emph{keyframe inspection} on candidate keyframes against the current prompt and memory, \emph{motion state grounding} to adapt the planned motion to the realized layout, and \emph{cross-shot propagation} to convert generated videos into memory updates and refined conditions for subsequent shots.

\subsubsection{Keyframe inspection}
Within shot $t$, the Analyzer first performs \emph{state consistency analysis} to evaluate each candidate keyframe against $q_t$ and $M_t$:
\begin{equation}
\begin{aligned}
  (a_t^{(k)}, q_t^{(k)}) &\gets \operatorname{Analyzer}_{\mathrm{inspect}} (I_t^{(k)}, M_t, q_t), \quad k = 1 \to K.
\end{aligned}
\end{equation}
Here, $a_t^{(k)}$ is the acceptance decision, and $q_t^{(k)}$ is the refined keyframe prompt returned upon rejection. Because the refinement names the specific state that is violated, regeneration is targeted rather than blind resampling. For example, when an underspecified prompt yields a back view that conflicts with a stored back-view state, the Analyzer specifies the intended pose and the Manager retrieves the matching reference before regeneration. If no candidate is accepted within $K$ rounds, the shot falls back to the best-matched candidate chosen by the Analyzer rather than stalling.

\subsubsection{Motion state grounding}
The accepted keyframe $I_t$ fixes the spatial layout that the planned narrative could not know in advance, so the Analyzer next performs \emph{motion state grounding}:
\begin{equation}
m_t \gets \operatorname{Analyzer}_{\mathrm{ground}}(I_t, m_t^0),
\end{equation}
rewriting the planned motion into one executable from that layout, e.g., adding the turn and walk that must precede a pickup when the target table ends up behind the protagonist. The Visual Generator then produces the target shot $V_t$ from keyframe $I_t$ and motion prompt $m_t$.

\subsubsection{Cross-shot propagation}
The generated video $V_t$ is the first evidence of how the planned prompts were actually realized. The Analyzer performs \emph{state transition analysis} to extract the changes that should affect subsequent generation, i.e.
\begin{equation}
\label{eq:post_video}
\begin{aligned}
  (U_t,q_{t+1}) &\gets \operatorname{Analyzer}_{\mathrm{propagate}} (V_t,M_t,m_t,q_{t+1}^0).
\end{aligned}
\end{equation}
The updates $U_t$ are written back through the four previously defined actions, while $q_{t+1}$ restates the planned $q_{t+1}^0$ in terms of the states now active in memory. For example, a newly revealed frontal face is appended as an additional view, an attire change is registered as a derived state and referenced by the next shot, and transient variations are ignored. Shot $t+1$ therefore starts from realized rather than merely planned visual history, which prevents early deviations from compounding across the sequence.

\begin{table*}[h]
  \centering
  \small
  \begin{tabular}{@{}l|c|c|cc|cc|c@{}}
    \toprule
    \multirow{2}{*}{Methods}
    & \multirow{2}{*}{\shortstack{Character\\Consistency}}
    & \multirow{2}{*}{\shortstack{Cross-Shot\\Similarity}}
    & \multicolumn{5}{c}{VLM-as-Judge} \\
    & & & Identity & Outfit & Prompt-Shot & Prompt-Video & Narrative  \\
    \midrule
    Wan2.2-T2V~\cite{wan2025wan}
    & 0.4489 & 0.5402 & 0.5727 & 0.4311 & 0.7274 & 0.3879 & 0.5402  \\
    \midrule
    StoryDiffusion~\cite{zhou2024storydiffusion}
    & 0.5083 & 0.5826 & 0.6705 & 0.1923 & 0.4835 & 0.1217 & 0.4943  \\
    HoloCine~\cite{meng2025holocine}
    & 0.5030 & 0.6105 & 0.9003 & 0.3807 & 0.4279 & 0.2732 & 0.6667  \\
    MultiShotMaster~\cite{wang2026multishotmaster}
    & 0.4794 & 0.5989 & 0.8907 & 0.6181 & 0.6213 & 0.4883 & 0.8111  \\
    StoryMem~\cite{zhang2025storymem}
    & 0.5739 & 0.6216 & 0.9611 & 0.7588 & 0.7252 & 0.4589 & 0.8333  \\
    \midrule
    Naive Frame Agent
    & 0.5787 & 0.6116 & 0.9678 & 0.7862 & 0.8069 & 0.6200 & 0.8542 \\
    \midrule
    Ours w/ Qwen3-VL
    & 0.5831 & 0.6156 & 0.9823 & 0.8054 & 0.8610 & 0.7950 & 0.8889  \\
    Ours
    & \textbf{0.6105}
    & \textbf{0.6389}
    & \textbf{0.9941}
    & \textbf{0.9238}
    & \textbf{0.8783}
    & \textbf{0.8400}
    & \textbf{0.9286}     \\
    \bottomrule
  \end{tabular}
  \caption{Comparison on the 30-story primary subset of StoryBench.}
  \label{tab:comparison}
\end{table*}

\begin{table*}[h]
  \centering
  \begin{tabular}{@{}l|c|c|cc|cc|c@{}}
    \toprule
    \multirow{2}{*}{Methods}
    & \multirow{2}{*}{\shortstack{Character\\Consistency}}
    & \multirow{2}{*}{\shortstack{Cross-Shot\\Similarity}}
    & \multicolumn{5}{c}{VLM-as-Judge} \\
    & & & Identity & Outfit & Prompt-Shot & Prompt-Video & Narrative  \\
    \midrule
    Ours w/o memory
    & 0.5415 & 0.5530 & 0.8262 & 0.5557 & 0.7960 & 0.7510 & 0.7021  \\
    Ours w/o synergy
    & 0.5908 & 0.5670 & 0.9841 & 0.8385 & 0.8360 & 0.7770 & 0.8582  \\
    \hfill- Best-of-$K$
    & 0.5914 & 0.5671 & 0.9838 & 0.8398 & 0.8372 & 0.7796 & 0.8411  \\
    Ours w/o refine
    & 0.5973 & 0.6089 & 0.9826 & 0.8454 & 0.8524 & 0.8204 & 0.9028  \\
    Ours w/o entity state
    & 0.5959 & 0.6170 & 0.9827 & 0.8433 & 0.8475 & 0.8430 & 0.9156 \\
    \midrule
    Ours
    & \textbf{0.6124}
    & \textbf{0.6323}
    & \textbf{0.9943}
    & \textbf{0.9348}
    & \textbf{0.8544}
    & \textbf{0.8700}
    & \textbf{0.9362}     \\
    \bottomrule
  \end{tabular}
  \caption{Ablation results on the full 50-story StoryBench.}
  \label{tab:ablation}
\end{table*}

\begin{table}[t]
  \centering
  \small
  \begin{tabular}{@{}lcccccc@{}}
    \toprule
    Metric & Wan & SDiff & Holo & MSM & SMem & Ours \\
    \midrule
    EC $\uparrow$ & 2.414 & 2.643 & 5.114 & 5.957 & 6.586 & \textbf{8.514} \\
    OC $\uparrow$ & 4.857 & 3.557 & 3.514 & 4.883 & 6.257 & \textbf{8.543} \\
    TVA $\uparrow$ & 5.700 & 4.571 & 3.643 & 4.971 & 5.786 & \textbf{7.800} \\
    \bottomrule
  \end{tabular}
  \caption{Human evaluation. Wan: Wan2.2; SDiff: StoryDiffusion+Wan2.2; Holo: HoloCine; MSM: MultiShotMaster; SMem: StoryMem. EC: entity consistency; OC: outfit correctness; TVA: text-video alignment.}
  \label{tab:human-eval}
\end{table}

\begin{figure*}[ht]
  \centering
  \includegraphics[width=0.99\linewidth]{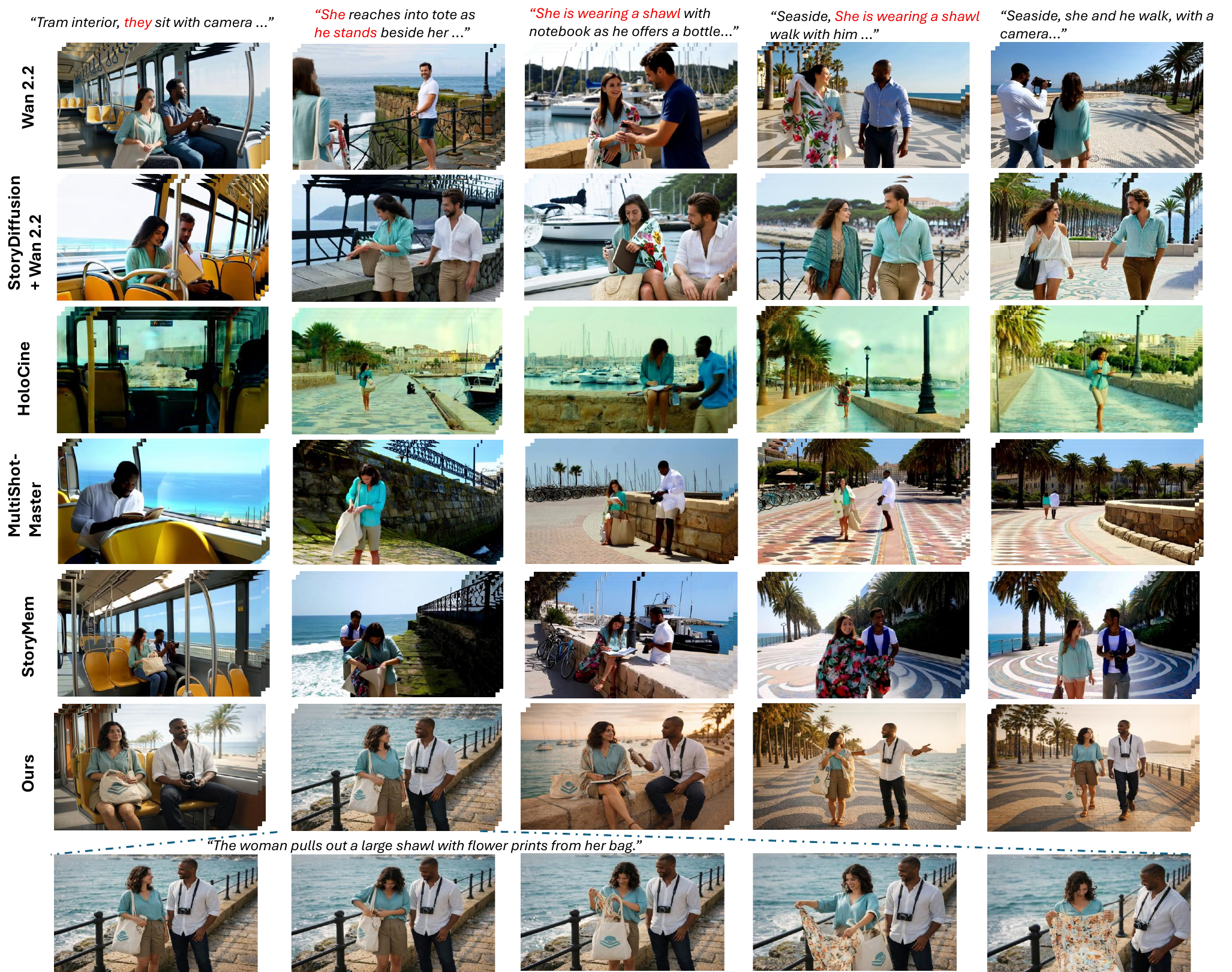}
  \caption{Qualitative comparison with other methods.}
  \label{fig:main comparison}
\end{figure*}

\section{Evaluation Benchmark}
\label{sec:storybench}

\subsection{Dataset Curation}
Existing video generation benchmarks can assess coherence over extended sequences but are less suited to precisely characterizing consistency under scene transitions, evolving entity states, and multi-event narrative progression. To address this gap, we introduce \textbf{StoryBench}, a multi-scene and multi-entity benchmark for story-driven multi-shot video generation. It contains 50 eight-shot stories across three subsets. The primary subset consists of 30 photorealistic human stories with two protagonists throughout all shots, providing a controlled setting for comparison with other methods. A second subset contains 10 photorealistic human stories with flexible character participation and delayed entity introduction. The remaining 10 stories adopt an anime style to broaden the visual styles. For each story, GPT-5~\cite{gpt5} generates an overall summary together with shot-level prompts, which are further decomposed into keyframe and motion prompts.

To evaluate consistency under entity-state evolution, we introduce controlled outfit changes as visual transitions, such as adding or removing a hat, changing a coat, or replacing a scarf. In the 30-story primary subset, one protagonist transitions from outfit~A to outfit~B and later reverts to outfit~A in fixed shots, testing whether a model can correctly apply, maintain, and reactivate appearance states while preserving identity. The additional 10 photorealistic stories use more flexible trajectories, where a protagonist may be absent or not clearly visible at the beginning and only appear later. Together, the three subsets support a targeted analysis of character consistency and narrative coherence under scene transitions, character-state evolution, and delayed entity introduction. Additional details are provided in the Appendix.

\subsection{Evaluation Metric}
\subsubsection{Character Consistency} We evaluate identity preservation within the same predefined appearance state. For each shot, we uniformly sample frames, localize the target character using GroundingDINO~\cite{liu2024grounding}, segment it with SAM~\cite{sam}, and extract DINOv2~\cite{oquab2024dinov2} features from the masked character regions. We then compute the mean pairwise cosine similarity among frames belonging to the same appearance state (e.g., outfit A or outfit B) and average the valid state-level scores with equal weight.

\subsubsection{Cross-Shot Similarity} We measure visual-semantic coherence across shots using the ViCLIP-based similarity metric~\cite{wang2024internvid} following~\cite{meng2025holocine}. For each predefined appearance state, we compute the average cosine similarity over all shot pairs within that state, and report the mean across states.

\subsubsection{VLM-as-Judge Evaluation}
To complement automatic metrics, we employ OpenAI o3~\cite{O3} as a VLM judge and evaluate several aspects: identity preservation, controlled appearance-state transitions, prompt realization, and narrative coherence. The judge follows a predefined rubric of binary ``yes''/``no'' questions, with applicability and aggregation weights specified in advance. 
\textbf{Identity} evaluates whether each character's persistent appearance attributes remain consistent across shots despite predefined state changes, including facial and head features, body shape, and other appearance details not intentionally modified.
\textbf{Outfit} evaluates whether outfit transitions occur at the intended shots, take the prescribed visual form, and remain consistent within each state.
\textbf{Prompt-Shot} evaluates whether each shot realizes the visual content, actions, and events specified by its shot prompt.
\textbf{Prompt-Video} evaluates whether the complete video realizes the intended story.
\textbf{Narrative} evaluates narrative progression, logical consistency, and motion continuity across shots.
More details are provided in the Appendix.

\section{Experiments}
\label{sec:experiments}

\begin{figure*}[ht]
  \centering
  \includegraphics[width=0.95\linewidth]{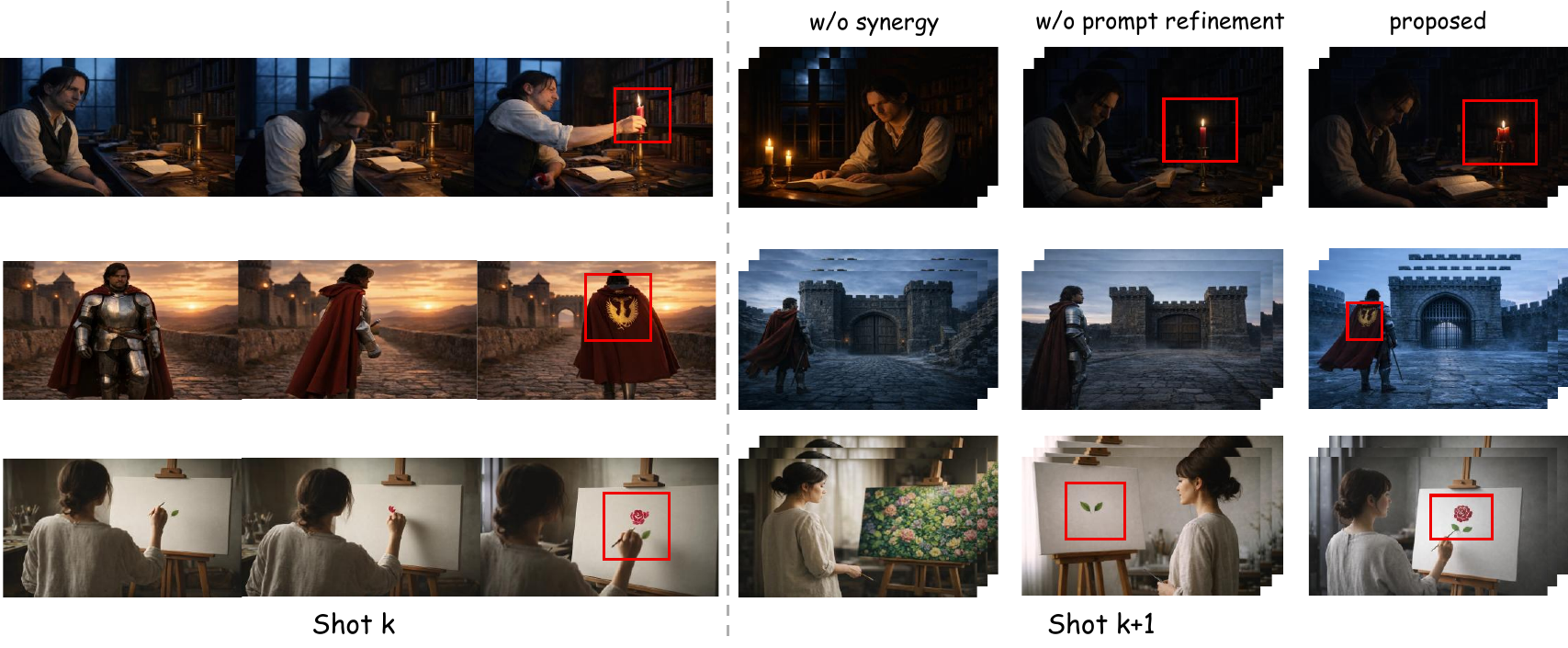}
  \caption{Qualitative ablation study. Upper (state evolution): candle consumption over time. Middle (view consistency): dorsal-view retrieval preserves identity across shots. Bottom (state awareness): retrieval of the \textit{finished} painting state.}
  \label{fig:ablation}
\end{figure*}

\subsection{Implementation Details}
We leverage GPT-Image-1.5~\cite{openai2025gptimage15} as the text-to-image and image-to-image generation model to generate the keyframe for each shot, and the state-of-the-art open-source model Wan2.2-I2V-A14B~\cite{wan2025wan} for video generation. We apply OpenAI o3 as the core VLM for the \emph{Visual State Analyzer} and \emph{State Memory Manager}. The resolution of all images and video clips is 832x480. Each shot is 5 seconds. The maximum rounds K is 3.

\subsection{Quantitative Comparison}
We compare our method with several representative baselines including (1) Wan2.2-T2V-14B, applied to each shot independently; (2) the two-stage StoryDiffusion~\cite{zhou2024storydiffusion} + Wan2.2-I2V-14B pipeline; and (3) recent multi-shot generation methods HoloCine~\cite{meng2025holocine}, MultiShotMaster~\cite{wang2026multishotmaster}, and StoryMem~\cite{zhang2025storymem}. We additionally construct a \emph{Naive Frame Agent}, which uses the same GPT-Image-1.5 and Wan2.2-I2V generators and storyboard prompts as ours but autoregressively retrieves previously generated frames through CLIP as image references, without state inspection, motion grounding, or propagation. This baseline isolates entity-centric evolving state management from the direct agentic use of frame-level visual history. We also report our framework with the open-sourced Qwen3-VL-32B~\cite{bai2025qwen3} as the Visual State Analyzer to examine its robustness to the analyzer backbone. Because some baselines impose constraints on visual style or character configuration, this comparison uses the 30-story photorealistic subset with persistent two-person casts for a controlled evaluation.

As shown in Table \ref{tab:comparison}, our method achieves the best performance across the board. It leads on both feature-level metrics, reaching 0.6105 in Character Consistency and 0.6389 in Cross-Shot Similarity, so character appearance and main composition are retained across shots. The advantage is more evident on the VLM-judged-based consistency. Identity score of 0.9941 demonstrates robust preservation of persistent appearance across the entire story, while an Outfit score of 0.9238 shows that the prescribed outfit changes are applied at the intended shots, maintained afterwards, and successfully reactivated later, which is exactly where the baselines degrade. Despite using the same generators, the Naive Frame Agent falls short on every metric; the gap is particularly large for state-dependent Outfit (0.7862 vs.\ 0.9238) and Prompt-Video (0.6200 vs.\ 0.8400), showing that retrieving raw frame history cannot replace explicit evolving-state management. Our method also achieves the highest Prompt-Shot, Prompt-Video and Narrative scores. The Qwen3-VL analyzer retains the performance gains under the same o3 evaluator, suggesting that the improvements are neither tied to a specific analyzer backbone nor solely explained by using the same model for analysis and evaluation.

\subsection{Human Evaluation}
We conduct a human study in Table~\ref{tab:human-eval} on 10 stories randomly sampled from the 30-story primary subset of StoryBench. Ten participants evaluate all 60 videos generated by five baselines and our method, rating entity consistency, outfit correctness, and text-video alignment on a 1--10 scale. This study complements automatic metrics by directly assessing whether characters remain recognizable across shots, outfit states are correctly maintained or transitioned, and generated visuals align with the prompts. Our method achieves the highest scores across all categories, consistent with the quantitative results.

\subsection{Qualitative Comparison}
Figure \ref{fig:main comparison} presents qualitative comparisons with five baselines. Compared with recent multi-shot methods MultiShotMaster and StoryMem, our method more reliably preserves character, prop, and background continuity, whereas the baselines often exhibit identity shifts, appearance inconsistencies, or structural drift in later shots.

Furthermore, our method demonstrates an exceptional capability to maintain the visual integrity of dynamic props. For instance, when a floral shawl is introduced in \emph{Shot 2}, its distinctive texture and pattern are accurately preserved throughout subsequent shots. In addition, our approach exhibits robust scene-level consistency. Given that the final two shots take place in the same seaside environment, our method models the background as a persistent, maintainable entity. This allows it to effectively preserve the global background and spatial layout across different temporal segments, ultimately producing a highly coherent multi-shot narrative.

\subsection{Ablation Studies}
Table~\ref{tab:ablation} evaluates the contribution of each component on the full 50-story StoryBench. \emph{Ours w/o memory} is an open-loop baseline using the same GPT-Image-1.5 and Wan2.2-I2V generators without Visual State Memory and Analyzer. \emph{Ours w/o synergy} retains memory state retrieval and registration but disables the Analyzer's triplet synergy. To distinguish the proposed synergy mechanism from a naive increase in sampling budget, we construct a \emph{Best-of-K} baseline on top of \emph{Ours w/o synergy}, which samples K keyframes from the original prompt and selects the best candidate. \emph{Ours w/o refine} uses the Analyzer but conditions image and video generation on the original prompts rather than Analyzer-refined prompts. \emph{Ours w/o entity state} keeps the Analyzer and closed loop unchanged but replaces entity-state memory entries with frames that the Analyzer judges informative or newly introduced. Later shots query this frame memory with the same VLM and receive at most four references, as in Ours. This isolates the memory representation while preserving the surrounding loop.

The fully sequential Ours w/o memory variant performs the worst. Introducing basic memory retrieval and update in Ours w/o synergy improves overall performance but remains substantially below Ours. Best-of-K yields almost no further improvement despite multiple generations, showing that our gains do not arise from simply sampling and selecting better images. With the Analyzer's memory maintenance enabled, Ours w/o refine achieves broad improvements, demonstrating the importance of observation-driven state updates and propagation. Nevertheless, it still underperforms Ours, indicating that prompt refinement provides additional gains by translating detected inconsistencies into targeted corrections. Ours w/o entity state stays close on prompt following and narrative quality, which benefit from the preserved closed loop, but remains below Ours on every metric. Its largest gap is on Outfit (0.8433 vs.\ 0.9348): frames capture a past appearance but not which state is currently active, so retrieval can return a plausible yet temporally stale appearance. This result shows that the closed loop alone is insufficient; binding observations to evolving entity states is essential for controllable long-horizon appearance transitions.

\textbf{Figure \ref{fig:ablation}} presents qualitative ablations of visual state synergy and prompt refinement.
\textbf{State Evolution (Case 1):} Without state observation, the model fails to track newly introduced candle. While memory alone retrieves the object, no prompt refinement ignores the temporal lapse from "evening" to "late night", yielding a pristine candle. Our full method correctly deduces physical consumption over time.
\textbf{View Consistency (Case 2):} Given an ambiguous prompt "knight gazing at a gate", raw text fails to bridge "knight" with the memory's "dorsal aspect", producing an inconsistent side view. Refinement explicitly targets the correct stored pose, preserving spatial and appearance consistency.
\textbf{State Awareness (Case 3):} Without explicit state management, the model either retrieves an outdated artwork state or hallucinates a different painting. Our framework successfully identifies the semantically evolved ``finished'' state, aligning the generation with narrative progress.

\section{Conclusion}
In this paper, we introduced \emph{EvoState}, an agentic framework that reformulates multi-shot video generation as observation-driven visual state evolution to overcome identity drift and compounding inconsistencies. A \emph{Visual State Analyzer} performs within-shot state alignment and cross-shot state propagation, while an entity-centric Visual State Memory separates persistent identity anchors from their evolved states. Together, these components continuously reconcile generated visual evidence with narrative intent through prompt refinement and state write-back. Experiments on our curated \emph{StoryBench} benchmark demonstrate substantial gains in identity preservation, state-transition correctness, and overall narrative quality compared with existing approaches.

\section{Ethics Statement}
Our work is purely a research project focusing on agentic framework to tackle identity drift and compounding inconsistencies in long-form multi-shot video generation. Currently, we have no plans to integrate this technology into any product or to make it publicly accessible. In future development, we will adhere strictly to Microsoft’s AI principles. Although we demonstrate results involving human faces in our research paper, the sole purpose is to showcase the capability of long-term identity coherence. We do not intend to generate content for misleading, deceptive, or harmful purposes. We strongly oppose any misuse of this technology to create misleading or harmful content involving real individuals. Additionally, we are interested in applying the key idea of this research to advance forgery detection techniques.

\bibliography{reference}

\twocolumn[
	\begin{center}
		{\LARGE\bfseries Supplementary Material}
	\end{center}
]

\appendix

\section{Computational Efficiency}
Table \ref{tab:cost} reports average per-shot computational cost. The low keyframe-regeneration frequency (0.1 per shot) keeps the state-evolution overhead modest; video generation remains the dominant cost. Latencies except for video generation are measured using API settings, while Wan2.2 video generation uses four 80GB A100 GPUs.

\begin{table}[h]
  \centering
  \small
  \begin{tabular}{@{}lrrrrrr@{}}
    \toprule
    Metric & $I_{\mathrm{regen}}$ & $A_{\mathrm{ins}}$ & $A_{\mathrm{gnd}}$ & $A_{\mathrm{prop}}$ & $I_{\mathrm{gen}}$ & $V_{\mathrm{gen}}$ \\
    \midrule
    Calls & 0.1 & 1.1 & 1.0 & 1.0 & 2.5 & 1.0 \\
    Time & 3.0 & 17.6 & 16.0 & 19.0 & 75.0 & 468.0 \\
    \bottomrule
  \end{tabular}
  \caption{Average per-shot cost. $A_{\mathrm{ins}}$, $A_{\mathrm{gnd}}$, and $A_{\mathrm{prop}}$ denote Analyzer calls for state inspection, motion grounding, and state propagation, respectively. Time is in seconds.}
  \label{tab:cost}
\end{table}

\section{Case Study}
\subsection{Cross-Shot Propagation}
\begin{figure*}[ht]
  \centering
  \includegraphics[width=0.95\linewidth]{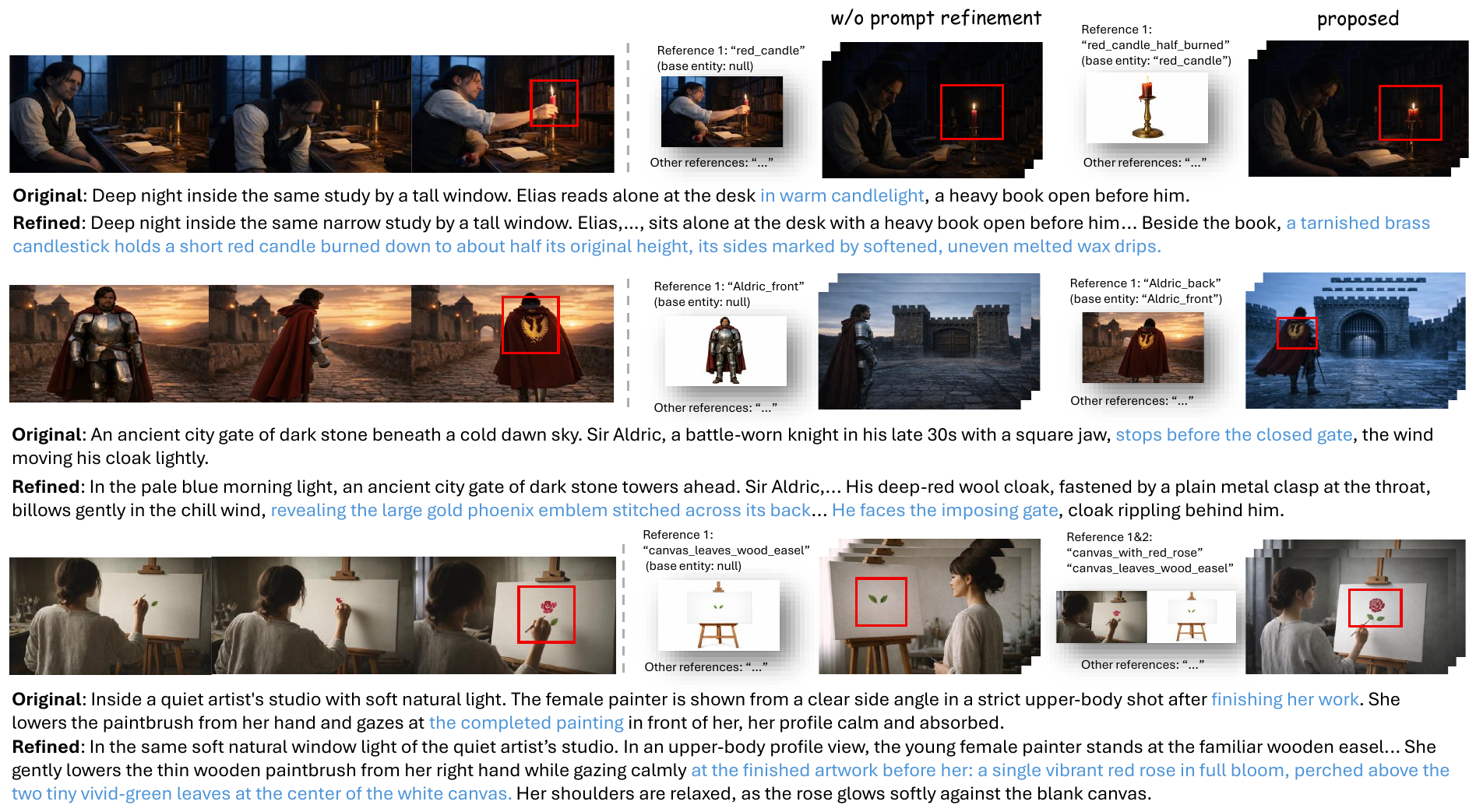}
  \caption{Original vs. refined keyframe prompts by inter-shot synergy, with their retrieved references.}
  \label{fig:t2i_refine}
\end{figure*}

As a supplement to the ablation study analysis, we provide the prompt refinement details in Figure \ref{fig:t2i_refine}. This process illustrates the inter-shot refinement mechanism by the Analyzer, where next-shot keyframe prompt is refined during text-visual-memory synergy. By bridging the underspecified textual intent with the accumulated visual history, this inter-shot refinement ensures that the generated keyframes seamlessly inherit established identities, explicit viewpoints, and temporally evolved object states (such as the burned-down candle and finished painting).

\subsection{Keyframe Inspect}
\begin{figure}[ht]
  \centering
  \includegraphics[width=1.0\columnwidth]{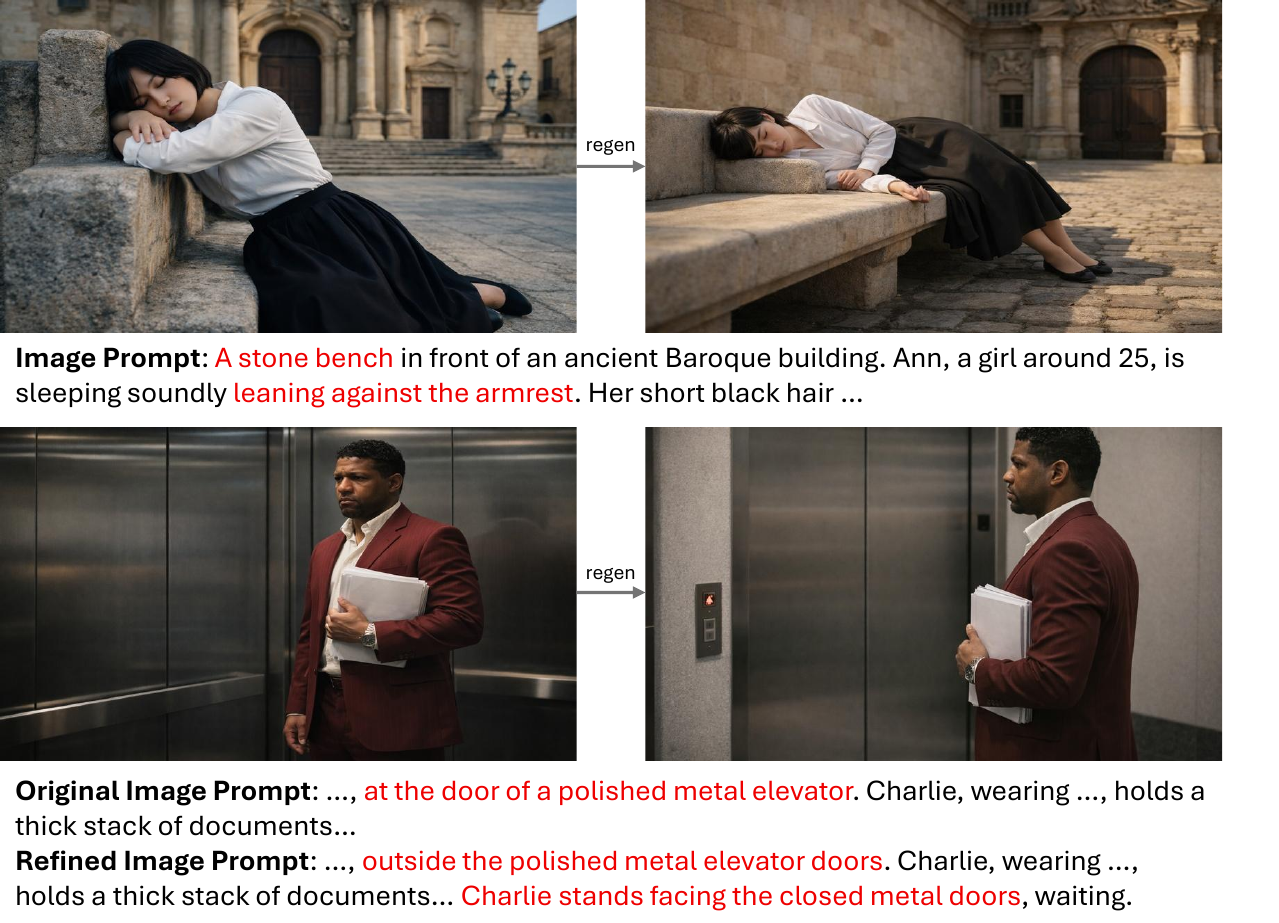}
  \caption{Examples of keyframe regeneration resolving prompt-following failures. Top: Direct regeneration corrects structural alignment without altering the prompt. Bottom: Regeneration with a refined prompt resolves spatial ambiguity.}
  \label{fig:regen}
\end{figure}

The keyframe regeneration mechanism acts as a critical corrective loop to ensure semantic alignment, as shown in Figure \ref{fig:regen}. This mechanism addresses prompt-following failures through two distinct approaches. In instances where the model stochastically misses spatial or structural details despite clear instructions—such as failing to accurately depict a character leaning against a bench armrest—direct regeneration without modifying the text often suffices to rectify the oversight (Figure \ref{fig:regen}, top). Conversely, when the initial prompt contains inherent ambiguities, such as the phrase "at the door" being misinterpreted as inside an elevator, the mechanism facilitates prompt refinement. By injecting a more explicit description, it eliminates vague phrasing to enforce precise spatial grounding (Figure \ref{fig:regen}, bottom). By leveraging the robust generative capabilities of GPT-Image-1.5, our pipeline minimizes the necessity for frequent interventions compared to smaller, less capable models. This design choice is strategic: while weaker models might show a larger marginal improvement from regeneration, the increase in inference latency and computational overhead would undermine the practical utility.

\subsection{Motion State Grounding}
\begin{figure*}[ht]
  \centering
  \includegraphics[width=0.95\linewidth]{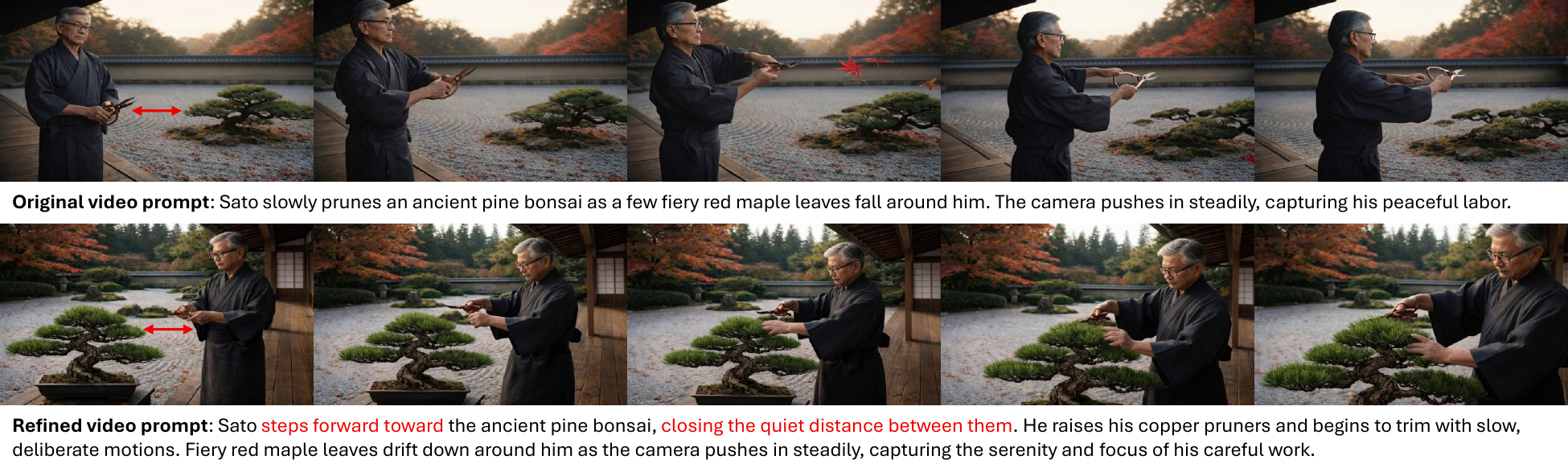}
  \caption{Original vs. refined video prompts by intra-shot synergy.}
  \label{fig:t2v_refine}
\end{figure*}

After the keyframe establishes the spatial layout and subject positioning, the video motion prompt should be refined to explicitly treat it as the initial frame and incorporate necessary transitional actions. Specifically in Figure \ref{fig:t2v_refine}, the description must account for any spatial gaps, e.g., if the man is initially positioned away from the bonsai, the prompt should include a forward step before pruning, so as to ensure coherent motion generation. Without such refinement, a generic prompt may lead to implausible actions (e.g., pruning gestures performed in midair) that fail to align with the established scene geometry.

\section{Benchmark and Metric}
\subsection{StoryBench Curation}
Every story is stored as eight shots, each carrying a keyframe prompt, a motion prompt and a duration, together with a specification that records the cast of each shot and the shot at which each prescribed outfit transition takes place. The three subsets instantiate this specification differently. In the 30-story primary subset both protagonists appear in all eight shots, and one of them wears outfit A in shots 1 to 3, puts on an additional garment at shot 4, keeps it through shot 6, and takes it off at shot 7 so that the last two shots return to outfit A. In \textit{story\_study} Sofia puts on a logo-printed cloak once the pair settles in the reading hall and folds it away before leaving. The added garment is always strongly patterned and visually unrelated to the base outfit, so a model that silently keeps or drops it is easy to detect.

The other two subsets relax this template in complementary directions. In the 10 photorealistic stories with flexible casting, the first character opens the story alone, the second character enters at shot 2 and is initially seen from behind or out of focus, the two share shots 3 to 5, and the second character then leaves permanently, after which the first character changes appearance around shot 6 and maintains the new appearance until the end without reverting. \textit{story\_paris} follows this pattern, where Amelie meets an elderly veteran by the Seine, continues alone after he departs, and wears a floral cape and a fedora in the closing shots. The 10 anime stories keep the full cast on screen in all eight shots and let one or two characters change in a later shot without recovery. The three templates therefore cover state evolution, state recovery, delayed entity introduction and entity exit. The complete storyboard prompts are submitted as a separate archive in the supplementary material.

\subsection{Details of Feature-Based Metrics}
Both metrics share one sampling scheme and one notion of an \emph{appearance state}. A story video is split into its $N=8$ shots and $4$ frames are sampled uniformly per shot. Each frame inherits a state label $g\in\{A,B\}$ from the benchmark specification, which fixes the position of the prescribed outfit transition; frames of shots where the character is not required to appear are skipped. Only within-state pairs are reported.

\subsubsection{Character Consistency}
For each sampled frame we run GroundingDINO with the character's appearance phrase as the text query (box threshold $0.3$, text threshold $0.25$) and keep the single highest-scoring box; frames with no detection are dropped. SAM segments that box, the mask is tightly cropped and all non-mask pixels are zeroed, so the descriptor sees the character alone and not the background or the co-occurring cast. The masked crop is encoded by DINOv2 ViT-B/14 into an $\ell_2$-normalized feature $f$. Write $\mathcal{F}_g = \{f_1,\dots,f_{n_g}\}$ for the $n_g$ features one character contributes in state $g$, pooled over all shots in which that state holds. The state score $C_g$ and the character score $C$ are
\begin{equation}
C_g = \binom{|\mathcal{F}_g|}{2}^{-1}\!\!\!\sum_{i<j,\,f_i,f_j\in\mathcal{F}_g}\!\!\! \langle f_i, f_j\rangle,
\qquad
C = \frac{1}{|\mathcal{G}|}\sum_{g\in\mathcal{G}} C_g,
\end{equation}
where $i$ and $j$ index $\mathcal{F}_g$ and the restriction $i<j$ keeps every unordered pair once while excluding the self-pairs that would each contribute a constant $1$. $C_g$ is the mean pairwise similarity inside state $g$. The outer average runs over $\mathcal{G}$, the set of states holding at least two valid segments, below which the pair count vanishes and $C_g$ is undefined. States are weighted equally, so a character who spends most shots in one state cannot let that state dominate the score. The reported number is the mean of $C$ over all characters of all stories, which weights characters rather than stories and therefore does not discount the harder multi-character stories.

\subsubsection{Cross-Shot Similarity}
This metric leaves the frames unmasked and measures coherence of the whole rendered image --- character, setting, and style together. Within each shot the sampled frames are grouped into maximal runs of consecutive frames sharing a state label, so a shot containing an on-screen transition contributes two runs instead of one. Each run is resampled to $8$ frames and encoded by ViCLIP into a single $\ell_2$-normalized clip embedding. Collecting the runs of one story in shot order into $\mathcal{R} = \{r_1,\dots,r_M\}$, we write $e_u$ for the embedding of run $r_u$ and $g_u\in\{A,B\}$ for the state label its frames share. The score of a story averages the cosine similarity over every pair of runs that carry the same label,
\begin{equation}
V = \frac{1}{|\mathcal{P}|}\sum_{(u,v)\in\mathcal{P}} \langle e_u, e_v\rangle,
\qquad
\mathcal{P} = \{(u,v): u<v,\ g_u = g_v\},
\end{equation}
so the indices $u$ and $v$ enumerate runs rather than shots or frames, and $u<v$ keeps every unordered pair exactly once. Two runs of the same shot always carry different labels, since a run is maximal, so each retained pair joins runs of two different shots and $V$ measures cross-shot rather than intra-shot coherence. The reported number is the mean of $V$ over stories. Since runs, not shots, are the unit of comparison, a shot in which the outfit changes contributes its pre-change and post-change halves to the states they actually belong to.

\subsection{Details of VLM-as-Judge}
The rubric consists of $31$ binary questions over four dimensions at two levels, listed in Tables \ref{tab:rubric_shot} and \ref{tab:rubric_video}. Shot-level questions inspect a single clip against its shot prompt and against its internal stability across frames. Video-level questions inspect the assembled story and ask whether characters, appearance states, and events hold together across shots. Narrative is defined at the video level only, since progression and cross-shot smoothness are not observable inside one clip.

Dimension-level questions are asked once per shot or per video, while identity and outfit questions are instantiated once per character. Applicability is derived automatically from the storyboard specification, and inapplicable questions are deleted from the prompt instead of being answered as ``N/A'', so that vacuous slots never dilute a score. Two answering rules keep the verdicts well-posed. Presence and adherence questions treat missing, unrecognizable, or unverifiable content as a failure, whereas stability questions count only visible and comparable evidence, so an item merely cropped out of frame is not read as drift. Identity questions ignore changing clothing and outfit questions ignore face and build, which prevents a single error from being charged in two dimensions. Every answer must cite the frame on which its evidence rests.

We use OpenAI o3 with deterministic decoding and issue one multimodal call per dimension per shot and per dimension per video, so that related questions share a common reading of the frames while an outfit failure cannot contaminate the identity verdict. Shot-level calls receive $8$ evenly spaced frames and video-level calls receive frames sampled at $1$ fps and capped at $40$ frames. Answers that cannot be parsed are dropped rather than counted as failures.

For a dimension $d$, the shot-level score of a story averages within each shot before averaging over shots, so that every shot counts equally regardless of how many characters it contains:
\begin{equation}
S_d^{\text{shot}} = \frac{1}{|\mathcal{S}|}\sum_{s\in\mathcal{S}} \frac{1}{|\mathcal{Q}_{d,s}|}\sum_{q\in\mathcal{Q}_{d,s}} \mathbb{1}[q = \text{yes}],
\end{equation}
where $\mathcal{Q}_{d,s}$ collects the applicable, answered slots of dimension $d$ in shot $s$. The video-level score $S_d^{\text{video}}$ averages the applicable video-level slots, and the two levels are combined with equal weight, $S_d = \tfrac12(S_d^{\text{shot}} + S_d^{\text{video}})$. \textbf{Identity} and \textbf{Outfit} in Tables \ref{tab:comparison} and \ref{tab:ablation} are reported in this combined form, and \textbf{Narrative} equals its video-level score. Only the prompt dimension is reported per level, as \textbf{Prompt-Shot} and \textbf{Prompt-Video}, since realizing a single shot and realizing the whole story fail for different reasons. All reported values are macro-averages over stories.

\begin{table*}[t]
  \centering
  \small
  \begin{tabular}{@{}llp{0.75\linewidth}@{}}
    \toprule
    ID & Aspect & Criterion (answered ``yes''/``no'') \\
    \midrule
    \multicolumn{3}{@{}l}{\textit{Prompt dimension}} \\
    P-S1 & Characters & All characters required by the shot are present, clearly visible, and visually match their identity description. \\
    P-S2 & Background & The setting and environment shown match the location the shot description specifies. \\
    P-S3 & Objects & All required key objects of the shot are actually visible.$^{\ddagger}$ \\
    P-S4 & Action & The action specified by the shot description actually happens on screen.$^{\ddagger}$ \\
    P-S5 & Style & The frames are rendered in the required visual style, i.e. photorealistic live-action rather than cartoon, anime, illustration, or 3D render. \\
    \midrule
    \multicolumn{3}{@{}l}{\textit{Identity dimension} (asked per character present in the shot)} \\
    I-S1 & Identity Stability & The character remains the same individual from the first to the last frame: no swap to another person, no identity morphing, no duplicate. \\
    I-S2 & Face Stability & Facial features (face shape, eyes, nose, mouth) stay stable and free of distortion across frames. \\
    I-S3 & Hair/Head Stability & Hairstyle, hair length, hair color, and facial hair stay stable across frames. \\
    I-S4 & Body Stability & Body build, height, and skin tone stay stable across frames. \\
    \midrule
    \multicolumn{3}{@{}l}{\textit{Outfit dimension} (asked per character present in the shot)} \\
    O-S1 & Initial State & At the beginning of the shot the character wears the annotated expected outfit (the source state in a change shot).$^{\ddagger}$ \\
    O-S2 & Required Change & The prescribed outfit change is actually performed on screen within this shot.$^{\dagger}$ \\
    O-S3 & Final State & At the end of the shot the character wears the annotated expected outfit (the target state in a change shot).$^{\ddagger}$ \\
    O-S4 & Within-shot Stability & Apart from the prescribed change, the outfit stays stable: no garment flickering in or out, no color or pattern drift between frames. \\
    O-S5 & Correct Direction & The change goes in the required direction, from the source state to the target state, rather than reversing or turning into some other garment.$^{\dagger}$ \\
    \bottomrule
  \end{tabular}
  \caption{Shot-level rubric. Each question is evaluated on $8$ evenly spaced frames of a single generated clip. $^{\dagger}$asked only in shots annotated as change shots; $^{\ddagger}$dropped when the annotation specifies no such requirement for the shot.}
  \label{tab:rubric_shot}
\end{table*}

\begin{table*}[t]
  \centering
  \small
  \begin{tabular}{@{}llp{0.75\linewidth}@{}}
    \toprule
    ID & Aspect & Criterion (answered ``yes''/``no'') \\
    \midrule
    \multicolumn{3}{@{}l}{\textit{Prompt dimension}} \\
    P-G1 & Event Coverage & The video shows every required event of the story at least once. \\
    P-G2 & Event Order & The required events appear in the order the specification prescribes. \\
    P-G3 & Scene Continuity & Judging the physical setting only, shots placed in the same location look like the same place (same room or site, same furniture and layout, same time of day), and the location changes only where the specification says it does. \\
    P-G4 & Prop Consistency & Every key object that recurs across shots keeps the same appearance --- color, shape, size, markings --- rather than being replaced by a visibly different object.$^{\ddagger}$ \\
    P-G5 & Cast Correctness & Considering the main characters only, the video shows exactly the specified cast: no additional prominent character, and no listed character rendered as two people at the same time. \\
    \midrule
    \multicolumn{3}{@{}l}{\textit{Identity dimension} (asked per character; dropped when the character appears in fewer than two shots)} \\
    I-G1 & Same Character & Across all shots where the character appears, it is recognizably the same individual, never replaced by a visibly different person. \\
    I-G2 & Face Consistency & Facial features are consistent wherever the character is required to appear. \\
    I-G3 & Hair/Head Consistency & Hairstyle, hair length, hair color, and facial hair are consistent wherever the character is required to appear. \\
    I-G4 & Appearance Consistency & Ignoring clothing entirely, skin tone, apparent age, and the distinguishing features named in the identity description are consistent each time. \\
    I-G5 & Relative Body Build & Body build and height are consistent, including relative to the other characters the subject appears alongside. \\
    \midrule
    \multicolumn{3}{@{}l}{\textit{Outfit dimension} (asked per character)} \\
    O-G1 & State-A Consistency & Across every shot where the character is expected in state A, that outfit looks the same: same garments, colors, and patterns.$^{\ddagger}$ \\
    O-G2 & State-B Consistency & The same criterion for state B.$^{\ddagger}$ \\
    O-G3 & State Inheritance & After the annotated change shot, the new outfit state persists unchanged into the following shots instead of silently reverting or drifting back.$^{\ddagger}$ \\
    O-G4 & Recovery Consistency & After the character changes back, the restored outfit matches the original state A exactly rather than being an altered version of it.$^{\ddagger}$ \\
    \midrule
    \multicolumn{3}{@{}l}{\textit{Narrative dimension}} \\
    N-G1 & Action Plausibility & Actions obey basic physical common sense: people and objects do not appear from nowhere or vanish, their number does not change without reason, nobody jumps to an implausible position, bodies and objects do not interpenetrate, and the visible result of an action matches the action itself. \\
    N-G2 & State Inheritance & When a character or object visibly changes state --- becoming injured, wet, dirty or dishevelled, picking up or putting down a carried item, an object being opened, cut, damaged or used up, or a lasting change to the setting --- that new state persists into the later shots instead of silently reverting. \\
    N-G3 & Story Progression & Every shot moves the story forward without unintended repetition: no action replayed that the specification requires only once, no narrative beat shown twice, and no shot that duplicates another. Repetition the specification explicitly requires does not count. \\
    \bottomrule
  \end{tabular}
  \caption{Video-level rubric. Each question is evaluated on frames sampled at $1$ fps across the whole story video, capped at $40$ frames. $^{\ddagger}$dropped when the annotation makes the question vacuous, e.g. no key object recurs across shots (P-G4), the character is never expected in that state (O-G1, O-G2), never changes outfit (O-G3), or never changes back (O-G4).}
  \label{tab:rubric_video}
\end{table*}

\section{Prompts}
\label{sec:app_prompts}
This section lists the system and user prompts of the two agentic modules, the \emph{Visual State Analyzer} and the \emph{State Memory Manager}. All of them read and write the same entity-state representation $e=(n,d,P,b)$ of the Visual State Memory and return strict JSON, so that every decision is directly executable rather than a free-form critique.

\subsection{Visual State Analyzer}
The Analyzer is a single VLM instantiated with three task-specific schemas that share this state representation.

\subsubsection{Keyframe inspection}
\label{sec:app_inspect}
This prompt implements $(a_t^{(k)},q_t^{(k)}) \gets \operatorname{Analyzer}_{\mathrm{inspect}}(I_t^{(k)}, M_t, q_t)$: the \texttt{decision} field carries the acceptance decision $a_t^{(k)}$ and \texttt{refined\_keyframe\_prompt} carries the refined condition $q_t^{(k)}$ returned upon rejection. Because the refinement names the state that is violated, the State Memory Manager can retrieve or register the matching reference before regeneration. When no candidate is accepted within $K$ rounds, the per-aspect alignment report is used to select the best-matched candidate from $C_t$ instead of stalling.

\begin{sideprompt}{SYSTEM PROMPT}
\textbf{Input:}
\begin{enumerate}[leftmargin=*, nosep]
\item Current generated candidate keyframe
\item Retrieved state reference images
\item Original keyframe prompt (source of truth)
\item Current keyframe prompt (possibly already refined)
\item Current Visual State Memory
\end{enumerate}

\textbf{EVALUATION FOCUS}
\begin{enumerate}[leftmargin=*, nosep]
    \item Main character presence and COUNT correctness (missing / duplicated main characters / deformed human body)
    \item Main character state alignment with the retrieved references (identity cues, outfit/accessories, silhouette, viewpoint)
    \item Environment/background alignment
\end{enumerate}

\textbf{DECISION RULES}
\begin{itemize}[leftmargin=*, nosep]
    \item If main character count is missing/duplicated/repeated incorrectly, return: "decision": "regen\_same\_desc" and DO NOT rewrite description.
    \item If there is a clear and material mismatch against the intended state --- identity, key wardrobe/props, the intended setting, or a viewpoint that conflicts with the retrieved state reference --- while core subjects are still correct, return: "decision": "regen\_refined\_desc" and provide a minimally adjusted description.
    \item If the image preserves the correct main subjects and overall scene intent, accept minor or non-critical differences in facial details, pose, clothing wrinkles, small accessory details, lighting, composition, or background details.
    \item Minor appearance drift or minor environment/background drift should NOT trigger regeneration.
    \item If sufficiently aligned overall, return: "decision": "accept".
\end{itemize}

\textbf{DESCRIPTION REWRITE CONSTRAINTS}
\begin{itemize}[leftmargin=*, nosep]
    \item Name the violated state explicitly (which character or prop, and which of its states is required) so that the required reference can be retrieved or registered before regeneration.
    \item Keep original intent and entities. Do NOT invent new characters, props, events, or story beats.
    \item Do NOT over-imagine. Only adjust emphasis and add small continuity details.
    \item Preserve proper nouns exactly.
    \item Prefer acceptance over regeneration when differences are subtle, ambiguous, or unlikely to materially improve with another generation.
\end{itemize}

\textbf{===== OUTPUT FORMAT =====} \\
\texttt{\{}\\
\texttt{~~"decision": "accept" | "regen\_same\_desc" | "regen\_refined\_desc",}\\
\texttt{~~"reasons": ["<string>", ...],}\\
\texttt{~~"alignment\_report": \{}\\
\texttt{~~~~"main\_character\_count": "ok" | "missing" | "duplicate" | "unclear",}\\
\texttt{~~~~"main\_character\_state": "ok" | "partial" | "mismatch" | "unclear",}\\
\texttt{~~~~"environment": "ok" | "partial" | "mismatch" | "unclear"}\\
\texttt{~~\},}\\
\texttt{~~"refined\_keyframe\_prompt": "<string>"}\\
\texttt{\}}\\

\textbf{Rules:}
\begin{itemize}[leftmargin=*, nosep]
\item "refined\_keyframe\_prompt" is required only when decision is "regen\_refined\_desc".
\item For other decisions, set "refined\_keyframe\_prompt" to "".
\item No markdown. No extra keys.
\end{itemize}
\end{sideprompt}

\begin{sideprompt}{USER PROMPT}
    \textbf{ORIGINAL KEYFRAME PROMPT:} \\
    {original\_keyframe\_prompt} \\

    \textbf{CURRENT KEYFRAME PROMPT:} \\
    {current\_keyframe\_prompt}  \\

     \textbf{VISUAL STATE MEMORY SUMMARY:} \\
    {current\_memory}   \\

    \textbf{IMAGE ORDER:}
    \begin{itemize}[leftmargin=*, nosep]
    \item Image 1: current generated candidate keyframe
    \item Images 2..N: retrieved state reference images
    \end{itemize}

    Review the alignment and return the strict JSON.

\end{sideprompt}

\subsubsection{Motion state grounding}
\label{sec:app_ground}
This prompt implements $m_t \gets \operatorname{Analyzer}_{\mathrm{ground}}(I_t, m_t^0)$, rewriting the planned motion into one executable from the layout that the accepted keyframe has fixed. The keyframe prompt is supplied only as a textual reference for what the image depicts; the image itself always prevails.

\begin{sideprompt}{SYSTEM PROMPT}
\textbf{Input:}
\begin{enumerate}[leftmargin=*, nosep]
    \item The accepted keyframe image (the first frame of the upcoming video shot)
    \item The keyframe prompt used to generate this keyframe
    \item The original motion prompt of this shot
\end{enumerate}

\textbf{Goal:} Rewrite the motion prompt so it starts from the exact visual content of the keyframe image, then describes the temporal progression. Use the keyframe prompt as a textual reference for what the keyframe depicts, but always prioritize the actual image content when they conflict.

    \textbf{START FROM THE KEYFRAME}
    \begin{itemize}[leftmargin=*, nosep]
        \item The keyframe image is the ground truth for the video's first frame.
        \item Your refined motion prompt \textbf{MUST} begin with a brief first-frame anchor: character appearance and positions kept concise and consistent with the keyframe prompt --- \textbf{do NOT} elaborate on clothing layers, accessories, background props, or architectural details.
        \item Insert whatever transitional movement the realized layout requires before the planned action can happen (e.g., a turn and a few steps toward a target that ended up behind the character).
    \end{itemize}

    \textbf{ACTION NARRATION}
    \begin{itemize}[leftmargin=*, nosep]
        \item The core of the description is \textbf{CHARACTER ACTIONS AND INTERACTIONS}: character-to-character interactions (dialogue, gestures, physical contact, eye contact, reactions) and character-to-prop interactions (picking up, using, holding, operating objects).
        \item These interactions \textbf{MUST} be the most prominent and detailed part of the description. They must \textbf{NOT} be buried under or diluted by appearance, environment, or atmosphere descriptions.
        \item Allocate the majority of the description's length and specificity to actions and interactions.
    \end{itemize}

    \textbf{ACTIONS MUST FOLLOW THE ORIGINAL MOTION PROMPT}
    \begin{itemize}[leftmargin=*, nosep]
        \item The temporal progression (what happens after the first frame) \textbf{MUST} follow the original motion prompt faithfully.
        \item \textbf{Do NOT} add, invent, or remove any main actions or plot beats.
        \item Minor wording adjustments for fluency are allowed, but the action sequence must remain the same.
    \end{itemize}

    \textbf{FORBIDDEN MOTIONS}
    \begin{itemize}[leftmargin=*, nosep]
        \item No close-to-wide or wide-to-close shot transitions within a single shot.
        \item No sudden perspective changes.
    \end{itemize}

\textbf{===== OUTPUT FORMAT =====} \\
\texttt{\{}\\
\texttt{~~"refined\_motion\_prompt": "<string>"}\\
\texttt{\}}\\
No markdown. No extra keys.
\end{sideprompt}

\begin{sideprompt}{USER PROMPT}
    \textbf{KEYFRAME PROMPT (keyframe reference):} \\
    {keyframe\_prompt} \\

    \textbf{ORIGINAL MOTION PROMPT:} \\
    {original\_motion\_prompt} \\

    \textbf{KEYFRAME IMAGE:} \\
    The image below is the accepted keyframe (first frame) for this shot.
    Rewrite the motion prompt starting from this keyframe's visual content, then follow the original action sequence.\\

    Generate the JSON output as specified in the system prompt. \\

\end{sideprompt}

\subsubsection{Cross-shot propagation}
\label{sec:app_propagate}
This prompt implements $(U_t,q_{t+1}) \gets \operatorname{Analyzer}_{\mathrm{propagate}}(V_t,M_t,m_t,q_{t+1}^0)$. Part A emits the observed state updates $U_t=\{(u_j,c_j)\}$, where every observation is paired with one of the four write-back actions, and Part B restates the planned condition $q_{t+1}^0$ in terms of the states now active in memory.

\begin{sideprompt}{SYSTEM PROMPT}
  You are a visual state analyst for a multi-shot video generation pipeline. \\
  You will be shown:
  \begin{enumerate}[leftmargin=*, nosep]
    \item Sampled frames from the GENERATED video of the current shot.
    \item The CURRENT Visual State Memory.
    \item The motion prompt of the current shot.
    \item The original keyframe prompt of the NEXT planned shot.
  \end{enumerate}

  \textbf{YOUR GOALS:}
  \begin{itemize}[leftmargin=*, nosep]
    \item \textbf{A) STATE WRITE-BACK:} Turn what the frames reveal into state updates, each paired with one write-back action.
    \item \textbf{B) NEXT SHOT KEYFRAME REFINEMENT:} Refine the next shot's keyframe prompt for visual continuity.
  \end{itemize}

  \textbf{===== PART A: STATE WRITE-BACK =====} \\
    \textbf{WRITE-BACK ACTIONS} \\
    For every observation you report, choose EXACTLY ONE action:
    \begin{itemize}[leftmargin=*, nosep]
      \item \texttt{append\_view}: the frames show a state that ALREADY exists in memory from a complementary angle, or more clearly than the stored reference, WITHOUT changing what that state is. Set \texttt{name} to the existing state and leave \texttt{identity\_anchor} null; the selected frame is attached to that entry as an additional reference. Example: a frontal face revealed for a character stored so far only from behind.
      \item \texttt{create\_state}: an entity in memory undergoes a continuity-critical and persistent visual change that must remain distinguishable from its earlier appearance --- a costume change, a layer or accessory added or removed that covers the body (coat, scarf, hat), a major injury, a prop broken or repainted, or a distinctly different side of the entity whose appearance obviously differs. Register a NEW state and set \texttt{identity\_anchor} to the exact name of the entity it evolves from.
      \item \texttt{create\_entity}: a plot-relevant character, signature prop, or signature background appears that NO entry in memory represents. Register it as a base state with \texttt{"identity\_anchor": null}.
      \item \texttt{ignore}: everything else. Transient poses, gestures, facial expressions, lighting / color grading / time-of-day drift, motion blur, temporary occlusion, minor handheld or nearby props, and anything you cannot confirm from the frames. This is the default choice.
    \end{itemize}

    \textbf{SCOPE \& PRIORITIES}
    \begin{itemize}[leftmargin=*, nosep]
      \item Strict Priority Order: 1) Main Characters (Highest); 2) Signature Props/Vehicles/Animals; 3) Signature Background/Location.
      \item Output Limit: at most 4 non-\texttt{ignore} updates per shot. Quality over quantity.
      \item If nothing qualifies, return empty: \texttt{"state\_updates": []}.
    \end{itemize}

    \textbf{FRAME SELECTION STRATEGY}
    \begin{itemize}[leftmargin=*, nosep]
      \item Frame Index: Specify the FRAME INDEX (0-indexed) that best captures the state.
      \item Clearest Frame Imperative: You MUST select the single clearest frame. Prioritize clear lighting (no face shadows), minimal motion blur, minimal occlusion, and maximal visibility of identity-critical details.
      \item Angle Flexibility: Frontal faces are NOT mandatory. Choose the viewpoint (front, side, 3/4, or back) that best displays critical visual evidence (e.g., distinctive silhouette, backpack, hairstyle, specific clothing layers, or inner/outer prop surfaces).
      \item Background Visibility: For background states, choose a frame where the background is most visible and least occluded by characters.
    \end{itemize}

    \textbf{DUPLICATE AVOIDANCE}
    \begin{itemize}[leftmargin=*, nosep]
      \item BEFORE any \texttt{create\_state} or \texttt{create\_entity}, compare carefully against the current memory. If the same identity/state/form is already stored, use \texttt{append\_view} or \texttt{ignore} instead of a renamed duplicate.
      \item An existing state with NO major form change MUST NOT be registered again.
      \item Do NOT split an entity and an inseparable accessory into two states (e.g., a character WEARING a distinctive hat).
    \end{itemize}

  \textbf{===== PART B: NEXT SHOT KEYFRAME REFINEMENT =====}\\
    \textbf{Basic Rules:} 1) Preserve the next shot's core intent and planned actions; 2) Restate it in terms of the states now active in memory: character states, persistent props; 3) MAINTAIN STYLE CONSISTENCY; 4) Use natural descriptive language (no numbering or special markers).

    \textbf{MULTI-SHOT INDEPENDENCE \& CONTINUITY LIMITS}
    \begin{itemize}[leftmargin=*, nosep]
      \item Multi-Shot Nature: Actions, plot events, and camera perspectives do NOT need to be continuous between consecutive shots.
      \item Selective Continuity: Inter-shot continuity strictly applies ONLY to main characters' identities, their currently active states, and specific persistent props and backgrounds.
      \item Prop Independence: Props do not need to appear continuously across all shots. Only include them if explicitly mentioned or logically required.
    \end{itemize}

    \textbf{ORIGINAL DESCRIPTION PRIORITY}
    \begin{itemize}[leftmargin=*, nosep]
      \item Source of Truth: The ORIGINAL next-shot descriptions are the definitive source for visual traits, plot details, and implicit compositions.
      \item Strict Feature Adherence: Do NOT alter specified appearance features (e.g., logos, prints, cartoon motifs) to match a generic "visual style".
      \item Preserve Perspective: Do NOT alter the implicit composition or camera position dictated by the original description (e.g., never change an implied frontal view to a side or back view).
    \end{itemize}

    \textbf{KEYFRAME COMPLETENESS}
    \begin{itemize}[leftmargin=*, nosep]
      \item Keyframe Anchor: Defines the FIRST FRAME. Must comprehensively detail character appearance (head-to-toe, layers, accessories), complete environment details, and character-environment relative positions.
      \item Focus: AVOID overly descriptive language on props, background and camera positions. Focus on the main characters and their key accessories.
      \item No State Alteration: DO NOT change the initial states of characters (e.g. pose, clothing state, facial expression) in the original keyframe prompt.
    \end{itemize}

  \textbf{===== OUTPUT FORMAT =====}\\
  \texttt{\{}\\
  \texttt{~~"state\_updates": [}\\
  \texttt{~~~~\{"action": "append\_view" | "create\_state"}\\
  \texttt{~~~~~~~~~~~~| "create\_entity" | "ignore",}\\
  \texttt{~~~~~"name": "<string>", "description": "<string>",}\\
  \texttt{~~~~~"frame\_index": <int>,}\\
  \texttt{~~~~~"identity\_anchor": "<string>" | null\},}\\
  \texttt{~~~~...}\\
  \texttt{~~],}\\
  \texttt{~~"refined\_next\_shot": \{}\\
  \texttt{~~~~"keyframe\_prompt": "<string>"}\\
  \texttt{~~\}}\\
  \texttt{\}}\\
  \texttt{description} is required for \texttt{create\_state} and \texttt{create\_entity}; for \texttt{create\_state}, copy the anchor's description and append a clause detailing the change. Set it to "" otherwise. No markdown. No extra keys.
\end{sideprompt}

\begin{sideprompt}{USER PROMPT}
    \textbf{CURRENT VISUAL STATE MEMORY:} \\
    {current\_memory} \\

    \textbf{CURRENT SHOT CONTEXT:} \\
    Motion prompt: {motion\_prompt} \\

    \textbf{NEXT SHOT ORIGINAL DESCRIPTIONS:} \\
    Keyframe prompt: {next\_keyframe\_prompt} \\

    \textbf{VIDEO FRAMES} \\
    The following frames are sampled from the CURRENT shot's generated video (0-indexed).
    Analyze these frames to:
    1) Report the observed state updates with their write-back actions.
    2) Refine the NEXT shot's keyframe description. \\

    Generate the JSON output as specified in the system prompt.

\end{sideprompt}

\subsection{State Memory Manager}
\label{sec:app_memory_prompt}
This prompt implements $(M_t,R_t) \gets \operatorname{ManageStateMemory}(M_t,q_t)$: \texttt{retrieved\_states} names the entries of $M_t$ that form the reference set $R_t$, while \texttt{new\_states} names the states the condition requires but memory does not yet hold, which the Visual Generator synthesizes and registers so that generation is never conditioned on an unresolved state. The manager is re-invoked after every condition refinement.

\begin{sideprompt}{SYSTEM PROMPT}
  Task: From the given keyframe prompt, extract the entity states the model should keep consistent.

  \textbf{PRIORITY ORDER}:
  \begin{enumerate}[leftmargin=*, nosep]
    \item Main characters (highest priority)
    \item Supporting characters, signature props/vehicles.
    \item Signature background (lowest priority).
  \end{enumerate}

  \textbf{EXTRACTION RULES}:
   \begin{enumerate}[leftmargin=*, nosep]
    \item STRICT LIMITS: Output at most 4 states. Do NOT invent entities not explicitly present or strongly implied by the shots. Prefer fewer, plot-central entities over minor ones.
    \item NAMING: Use stable, reusable names. Reuse an existing state name ONLY if its visual form, state, and viewpoint semantics perfectly match the current shot.
    \item DESCRIPTIONS: Each description must be a comprehensive, standalone prompt for image generation. For characters: detail head-to-toe clothing layers and accessories (do NOT force frontal faces; emphasize the most continuity-informative angle and silhouette). For props: specify shape, material, color, and markings.
   \end{enumerate}

  \textbf{EVOLVED STATE POLICY (Angles, States \& Accessories)}: \\
  When an existing entity requires a continuity-critical visual change, you must request it as a separate state bound to that entity's identity anchor. \\
  TRIGGERS FOR EVOLUTION: \\
  Create an evolved state ONLY under these three conditions:
  \begin{itemize}[leftmargin=*, nosep]
    \item Viewpoint/Angle Shifts: The shot requires tracking a distinctly different visual side of the entity which OBVIOUSLY brings appearance changes (e.g., front vs. back of a character's outfit (specific visual details), inner vs. outer face of a shield).
    \item State Changes: The entity undergoes obvious physical modification (e.g., severely injured, aged, transformed, or a prop gets broken/painted).
    \item Accessory/Costume Drifts: A character adds or removes continuity-critical clothing layers or accessories that significantly alter their silhouette or cover parts of the body (e.g., donning a heavy coat, scarf, shawl, or hat).
  \end{itemize}

  \textbf{FORMATTING EVOLVED STATES}:
  \begin{itemize}[leftmargin=*, nosep]
    \item Naming Convention: Use [AnchorName]\_[State] (e.g., "Mia\_back\_outfit", "Kaito\_injured", "shield\_inner\_face").
    \item Linkage: You MUST set the identity\_anchor key to the exact name of the entity this state evolves from.
    \item Description Rule: Do NOT write a new description from scratch. Copy the anchor's exact description, then append a clear, specific clause detailing the change or angle (e.g., "The same outfit as Mia\_front\_outfit, but viewed from the back, showing the red backpack and hair ribbons.").
  \end{itemize}

  \textbf{SIGNATURE BACKGROUND STATE RULES}:
  \begin{itemize}[leftmargin=*, nosep]
    \item Only output ONE background state (EMPTY scene: no characters and important props) at most.
    \item The description MUST be specific and stable (layout, key structures, dominant materials/colors, time-of-day/weather) so it can be reused.
    \item Naming convention: use a stable name like "bg\_apartment\_living\_room".
  \end{itemize}

  \textbf{DUPLICATE AVOIDANCE RULE}:
  \begin{itemize}[leftmargin=*, nosep]
    \item Before proposing ANY new\_states, you MUST compare against KNOWN STATES carefully.
    \item If a KNOWN STATE already represents the same identity/state/form, you MUST output it in retrieved\_states instead of creating a new name.
    \item This applies especially to evolved states already saved in memory. Do NOT create a synonymous new name for an already-known evolved state.
    \item When an existing front/back state already matches the needed form, reuse it. Do NOT create another front/back variant with a different name.
    \item DO NOT treat an accessory and a character with the accessory as two separate states if they are visually inseparable. (e.g., a character WEARING a distinctive hat should not be split into "character" + "hat".).
    \item The evolved states written back from previous shots can be considered higher priority. DO NOT create a new state if the same evolved state already exists in memory.
  \end{itemize}

  \textbf{STRICT NON-STATE EXCLUSIONS}:
  \begin{itemize}[leftmargin=*, nosep]
    \item Do NOT create a new state for temporary actions, poses, gestures, facial expressions, or camera-dependent moments.
    \item Do NOT create a new state for descriptions such as running, walking, turning, looking back, raising a hand, sitting, standing, kneeling, smiling, shouting, or other transient motion states.
    \item Do NOT create a new state for minor nearby props or temporary handheld items unless they are clearly plot-central, recurring, and visually important across shots.
    \item Examples of items that are usually NOT states unless clearly central and recurring: cups, plates, loose papers, random phones, small tools, chairs, table clutter, background decorations.
    \item If uncertain whether something should be a new state, prefer reusing an existing state or output fewer states.
  \end{itemize}

  \textbf{===== OUTPUT FORMAT =====}\\
  \texttt{\{}\\
  \texttt{~~"retrieved\_states": [}\\
  \texttt{~~~~\{"name": "<string>"\}, ...}\\
  \texttt{~~],}\\
  \texttt{~~"new\_states": [}\\
  \texttt{~~~~\{"name": "<string>", "description": "<string>",}\\
  \texttt{~~~~~"identity\_anchor": "<string>" | null\}, ...}\\
  \texttt{~~]}\\
  \texttt{\}}\\
  No extra keys. No markdown.
\end{sideprompt}

\begin{sideprompt}{USER PROMPT}
KNOWN STATES: \\
{current\_memory} \\

KEYFRAME PROMPT:\\
{keyframe\_prompt} \\

Return the states needed for THIS shot. Prioritize characters and props. Be conservative on evolving states.
\end{sideprompt}

\end{document}